\pdfoutput=1
\documentclass[a4paper]{article}
\usepackage{nopageno}
\usepackage{graphicx}
\usepackage{gensymb}
\usepackage{amsmath}
\usepackage{amsfonts}
\usepackage{amssymb}
\usepackage{amsthm}
\usepackage{hyperref,xspace,subfigure,fullpage}
\usepackage[footnotesize]{caption}

\title{Invariant Spectral Hashing of Image Saliency Graph}

\author{Maxime Taquet, Laurent Jacques, Christophe De Vleeschouwer and Beno\^{i}t Macq\\[2mm]
Information and Communication Technologies, Electronics and
Applied Mathematics\\
Universit{\'e} catholique de Louvain,
Belgium.
\footnotetext{$^*\,$MT, LJ and CDV thank the Belgian National Science Foundation
(F.R.S.-FNRS) for its financial support.}\\
  \texttt{maxime.taquet@uclouvain.be},
  \texttt{laurent.jacques@uclouvain.be} }

\newcommand{\ie}{\emph{i.e.}, }
\newcommand{\eg}{\emph{e.g.}, }
\newcommand{\etal}{\emph{et al.}\xspace}
\newcommand{\cl}{\mathcal}
\newcommand{\bs}{\boldsymbol}
\newcommand{\Ebb}{\mathbb{E}}

\newcommand{\Rbb}{\mathbb{R}}

\newcommand{\ud}{{\rm d}}

\newcommand{\inv}[1]{\frac{1}{#1}}

\DeclareMathOperator{\diam}{diam}
\newtheorem*{definition}{Definition}
\newcommand{\sq}{\vspace{0mm}}

\begin{document}

\maketitle

\sq\sq\sq\sq
\begin{abstract}
  Image hashing is the process of associating a short vector of bits
  to an image. The resulting summaries are useful in many applications
  including image indexing, image authentication and pattern
  recognition. These hashes need to be invariant under transformations
  of the image that result in similar visual content, but should
  drastically differ for conceptually distinct contents. This paper
  proposes an image hashing method that is invariant under rotation,
  scaling and translation of the image.  The gist of our approach
  relies on the geometric characterization of salient point
  distribution in the image. This is achieved by the definition of a
  \emph{saliency graph} connecting these points jointly with an image
  intensity function on the graph nodes. An invariant hash is then
  obtained by considering the spectrum of this function in the
  eigenvector basis of the Laplacian graph, that is, its graph Fourier
  transform. Interestingly, this spectrum is invariant under any
  relabeling of the graph nodes. The graph reveals geometric
  information of the image, making the hash robust to image
  transformation, yet distinct for different visual content.  The
  efficiency of the proposed method is assessed on a
  set of MRI 2-D slices and on a database of faces.

\noindent {\bf Keyworks} Invariant Hashing, Geometrical Invariant, Spectral
  Graph, Salient Points.
\end{abstract}

\sq
\section{Introduction}
\sq
\label{sec:introduction}

Summarizing images by much shorter sets of bits is of strong interest
for many different image processing applications. The summaries, or
\emph{hashes}, can be used as content identification to efficiently query
images in a database. In shape matching, hashes can represent patterns
of interest in order to find corresponding patterns~\cite{gal06}. Key
dependent hashes can also be used to authenticate images and ensure
their integrity~\cite{monga06}.

Image hashing is usually performed in two
steps~\cite{mihcak01}. First, an intermediate hash is produced by
extracting a representative set of parameters from the image. Second,
this intermediate hash is quantized by means of vector quantization in
order to increase its robustness while reducing its effective bit
size. These two steps are independent and this paper focuses on the
first step, namely the production of an intermediate hash.

One main challenge in image hashing is the robustness of the summary
with respect to image transformations preserving the visual
content. This robustness should be ensured while preserving the
ability to distinguish distinct visual contents. Different authors
have addressed this problem by proposing hashing methods based on
image features~\cite{lamdan88, venkatesan00, mihcak01,monga06}.
In~\cite{lamdan88}, the hash is produced by locating features points
and recording their relative coordinates in the orthonormal frame
defined by two of them. The operation is repeated for all possible
pairs of features points. Their approach is robust to global
transformations and partial occlusion. However, it is limited to
relatively simple patterns as they require the storage of many
coordinates. In \cite{venkatesan00}, the wavelet transform of the
image is computed and each subband is tiled into rectangles. The
variances or mean value of the intensities is computed for each
rectangle and concatenated to produce the intermediate hash. The
method presented in~\cite{mihcak01}, uses an iterative region growing
in the coarse subband of the discrete wavelet transform and simply
records the location of the salient points as the intermediate hash
value. Clearly, these two last methods achieve relatively poor results
for large image rotation and scaling, since they strongly depend on
the order of selection of the features points. In~\cite{monga06},
features points are extracted as the locations for which an
end-stopped wavelet transform is maximized. The recorded hash function
is then the normalized histogram of the corresponding wavelet
coefficients. Although the features histogram is more invariant under
rotation and scaling, it still cannot ensure invariance of the hash
values under large rotation. Besides, Kokiopouou \etal
\cite{kokiopoulou09} have recently developed a metric between pattern
transformation manifolds and achieved excellent results in terms of
rotation and scale invariance. However, their approach is not applied
for image hashing and uses orthogonal matching pursuit which is
computationally cumbersome. The lack of robustness under large
rotation for most common image hashing methods has been recently
identified in~\cite{yang09}. The author therefore proposes a novel
hashing approach whose efficiency does not depend on the rotation
angle. His approach is based on mean luminance information over
image sectors. Although more robust to large rotation, his method is
not robust under scaling of the image.

This paper proposes a hashing method that is, by construction,
invariant under rotation of any angle and under scaling up to
interpolation that preserves the significant structures. The presented
hash function is built in two steps. First, given a simple salient
point detector (Sec.~\ref{sec:smooth-harr-detect}), a smoothed version
of the Harris corner detector \cite{harris1988combined}, a
\emph{saliency graph} is constructed
(Sec.~\ref{sec:graph-definition}). This structure is a (weighted)
undirected graph connecting geographically close salient
points. Second, the graph Fourier transform of a function defined on
the graph, that is, its spectrum in the Laplacian graph eigenvector
basis, is computed (Sec.~\ref{sec:spectral-hashing}). The use of this
graph Fourier transform makes the hash independent of the salient
point selection order. Moreover, in order to ensure invariance under
transformations of the image, both the feature points selection and
the definition of the function need to be invariant. A particular
attention is therefore brought to the invariance of these last two
elements. Sec.~\ref{sec:results} presents finally the results of the
method applied on the Brainweb database of brain MRI
images~\cite{kwan99} and on the ORL Database of
Faces~\cite{samaria94}.

\sq
\section{Saliency Graph}
\sq
\label{sec:saliency-graph}
 

Our image hashing method relies on the definition of a \emph{saliency
  graph} built from particular salient points and from a certain
geographical connectivity between them. This graph will be used in
Sec.~\ref{sec:spectral-hashing} for summarizing functions of its node
locations in a geometrically consistent way.  Hereafter, we first
explain the method used to detect salient points, and then describe
how the graph can be generated from them.

\sq
\subsection{Smoothed Harris Corner Detector}
\label{sec:smooth-harr-detect}

We define our salient points as the intensity \emph{corners}
discovered by a smoothed Harris detector
\cite{harris1988combined,lindeberg1998feature}. These specific points
are indeed preserved under the transformation of interest, that is,
under image rotation, translation and scaling. Let us describe briefly
this method while insisting on the properties of interest for our
approach.

The smoothed Harris detector aims at detecting corners on the
principle that around these points the local intensity gradient
strongly varies. Mathematically, given a continuous model $I(\bs x)$
of the image intensity at location $\bs x=(x,y)\in\Rbb^2$, the
smoothed Harris corner detector at scales $0<\sigma<\tau$ uses the
matrix field
\begin{equation}
  \label{eq:SHD-def}
  J^{(\sigma,\tau)}(\bs x)\ =\ \int_{\Rbb^2}\ [\bs \nabla I^{(\sigma)}\bs\nabla^T
  I^{(\sigma)}](\bs x')\ g^{(\tau)}(\bs x-\bs x')\ \ud^2 \bs x'\ \in\Rbb^{2\times 2},
\end{equation}
where $g^{(\sigma)}$ is the Gaussian kernel of variance $\sigma^2$,
$I^{(\sigma)}(\bs x)=[I\ast g^{(\sigma)}](\bs x)$ is the smoothed copy
of $I$ and $\bs\nabla$ stands for the 2-D gradient operator. In other
words, since the rank 1 matrix $[\bs\nabla I^{(\sigma)}\bs\nabla^T
I^{(\sigma)}](\bs x)$ has for eigenvector the gradient $\bs\nabla
I^{(\sigma)}(\bs x)$ itself, the matrix $J^{(\sigma,\tau)}(\bs x)$
studies the variability of this vector in a neighborhood of $\bs x$
determined by the window $g^{(\tau)}$. In this paper, we arbitrarily
set $\tau=3\sigma$ in order to have a neighborhood with enough
gradient variations, and we give up hereafter the extra parameter
$\tau$ in the notations.

Since the Gaussian kernel is isotropic, $J^{(\sigma)}(\bs x)$ is
invariant under rotation. If $I(\bs x) \to I(R^{-1}_\theta \bs x)$ for the common ${2\!\times\!2}$ rotation
matrix $R_\theta$ of angle $0\leq \theta < 2\pi$, we show easily that
$J^{(\sigma)}(\bs x) \to R_{\theta}\,J^{(\sigma)}\big(R^{-1}_\theta
\bs x\big)\,R^{T}_{\theta}$.  In particular, the eigenvalues of
$J^{(\sigma)}$ remain unchanged under image rotation.  Moreover, if
the image undergoes a rescaling $I(\bs x)\to I(\bs x/\xi)$ for
$\xi>0$, $J^{(\sigma)}(\bs x) \to c\,
J^{(\frac{\sigma}{\xi})}(\bs x/\xi)$ for some spatially invariant
$c>0$, which links eigenvalues across scales. Under a more realistic
discrete model of the image intensity $I$ where $\bs x$ is taken on a
pixel grid, these invariances remain approximatively true as long as
$\sigma$ is larger than few multiples of the pixel size.

The smoothed Harris corner detector proceeds by analyzing the two eigenvalues
$\zeta_1(\bs x) < \zeta_2(\bs x)$ of $J^{(\sigma)}(\bs
x)$. Indeed, on image corners, both eigenvalues are strong and
positive \cite{harris1988combined,lindeberg1998feature}, while along
straight edges, $0 \simeq \zeta_1 < \zeta_2$. This
characterization is observed through the \emph{cornerness} of $I$, that
is,
$$
\cl C^{(\sigma)}(\bs x)\ =\ \det J^{(\sigma)} - \kappa\,{\rm
tr}\big(J^{(\sigma)}\big)^2\ =\ \zeta_1\zeta_2
-\kappa\,\big(\zeta_1 + \zeta_2\big)^2,
$$
for some $\kappa>0$ (typically set to $\kappa=0.04$). Corners are then
defined as the local maxima of the cornerness (as illustrated on
Fig.~\ref{fig:harris-example}), that is,
\begin{equation}
  \label{eq:corner-set-def}
  \cl V^{(\sigma)}\ =\ \{\bs x\,:\ \cl C^{(\sigma)}(\bs x)\ \textrm{is locally maximum}\}.
\sq
\end{equation}

\paragraph{Corner points invariance:} 
The elements of $\cl V^{(\sigma)}$ inherit the geometrical invariance
of $J^{(\sigma)}$ described above. This fact is obvious for
translation and rotation. For image scaling, if $I(\bs x)\to I(\bs
x/\xi)$, $J^{(\sigma)}(\bs x) \to c\, J^{(\sigma/\xi)}(\bs x/\xi)$ for
some $c>0$ independent of $\bs x$, and $\cl V^{(\sigma)} \to \xi\, \cl
V^{(\sigma/\xi)} = \{\xi \bs x: \bs x\in \cl
V^{(\sigma/\xi)}\}$ since $\cl C^{(\sigma)}(\bs x) \to c^2\,\cl
C^{(\sigma/\xi)}(\bs x/\xi)$.

\sq
\paragraph{Size of $\cl V^{(\sigma)}$:} 
Generally, the size of $\cl V^{(\sigma)}$ is controlled by
thresholding small values of $\cl C^{(\sigma)}$ in
(\ref{eq:corner-set-def}). In this work we prefer an adaptive
formulation where we keep only a fixed number of the strongest local
maxima in the cornerness. This will be useful latter to control the
size of the graph defined from $\cl V^{(\sigma)}$.

\sq
\paragraph{Choice of $\sigma$ and scale invariance:} 
In order to define an object-dependent smoothing scale $\sigma^*$, we
first compute the set $\cl V^{(\sigma_0)}$ with a minimal scale
$\sigma_0$ set to few pixels. This first point set is voluntary
dense. However, we can compute its \emph{diameter} $\diam(\cl
V^{(\sigma_0)})$, with $\diam(\cl A)=\max_{x,x'\in\cl A}{\rm dist}(\bs
x, \bs x')$ for any set of pixels $\cl A\subset\Rbb^2$. If the image
contains only one object\footnote{The conclusion describes a possible
  generalization for images with several objects on a smooth
  background.}, this diameter is close to the diameter of the object
itself. Therefore, by setting in a second round the object-dependent
scale $\sigma^* = \rho\, \diam(\cl V^{(\sigma_0)}) > \sigma_0$, for
some $0<\rho<1$, the aforementioned scale invariance of the corner set
makes $\cl V^{(\sigma)}$ scale invariant\footnote{Of course, this
  holds only for scaling factor compatible with the image
  sampling.}. In particular, $(\diam \cl V^{(\sigma^*)})^{-1}\ \cl
V^{(\sigma^*)}$ remains identical if $I(\bs x)\to I(\bs x/\xi)$. With
this procedure in hand and setting arbitrarily $\rho=0.025$ for the
typical application of Sec.~\ref{sec:results}, the resulting corner
set $\cl V^{(\sigma^*)}$ is simply written~$\cl V$.

\sq
\subsection{Graph definition}
\label{sec:graph-definition}

In order to reveal geometric information of the image $I$, a graph can
be built upon the detected salient points. A ``Saliency Graph'' is
therefore defined as the undirected graph $\cl G=\cl G(I)=(\cl V, \cl
W)$ connecting the corner points $\cl V=\{\bs c_i: 1\leq i \leq N_c\}$
through the definition of the \emph{connectivity} matrix $\cl
W\in\Rbb^{N_c\times N_c}$. In other words, given the diameter
$d^*=\diam(\cl G)=\diam(\cl V)$ and a certain radius $r>0$ defined
later, the connection between $\bs c_i$ and $\bs c_j$ is weighted by
$(\cl W)_{ij}$ (a zero weight meaning no connection) and the full
matrix reads
$$
(\cl W)_{ij}\ =\ 
\begin{cases}
\exp(-\inv{2\,r^2\, (d^*)^2}\,\|\bs c_i - \bs c_j\|^2), &\text{if $i\neq
j$ and $\|\bs c_i - \bs
c_j\| \leq 3\,r d^*$},\\
0,&\text{else},
\end{cases}
$$
where the value 3 ensures that the exponential is set to 0 if it falls
below 1.1\% of its peak value.

This connectivity choice is motivated by the wish to converge towards
the true space geometry when the number of nodes increases
\cite{singer2006gml}. In particular, since the node set discretizes
the planar domain, the following graph Laplacian
$$
\Delta\ =\ \cl E - \cl W\ \in\,\Rbb^{N_c\times
N_c},\quad\text{with}\  \cl E_{ij} = \big(\textstyle\sum_k \cl W_{ik}\big)\,\delta_{ij},
$$
tends to the continuous planar Laplacian if $N_c \to \infty$. Notice
that, whatever $\cl G$, the vector of ones $\bs 1\in\Rbb^{N_c}$ is
such that $\Delta \bs 1 = 0$, that is, $\bs 1$ is an eigenvector of zero
eigenvalue.

The purpose of the Saliency Graph $\cl G$ is to capture the
distribution geometry of the salient points.  The definition of the
connectivity $\cl W$ is therefore of paramount
importance. Interestingly, the radius $r$ weights the impact of the
geometry: if $r\to+\infty$ or if $r\to 0^+$, all the nodes are either
inter-connected with unit weight (\emph{complete} graph), or fully
disconnected ($\cl W=0$). In such limit cases, knowledge about the
salient point distribution is completely lost.  The radius $r$ should
therefore be selected carefully between these two extreme cases.

\sq
\section{Invariant Spectral Hashing}
\sq
\label{sec:spectral-hashing}


Spectral Graph theory \cite{chung97} studies the property of a graph
through the spectrum of its Laplacian operator. In particular, the
$N_c$ Laplacian eigenvectors 
$$
\cl B = \{\bs v_j\in\Rbb^{N_c}: 1\leq j\leq N_c,\ \Delta\,\bs v_j =
\lambda_j \bs v_j\,\},\quad\text{with}\ \bs v_1=\bs 1,\ \lambda_1=0,\ \lambda_{j}\leq\lambda_{j+1},
$$
constitute an orthonormal basis of $\Rbb^{N_c}$, that is, a basis any
function $\bs f\in\Rbb^{N_c}$ defined on the graph nodes. This basis
$\cl B$ can be alternatively represented as the matrix $\cl B = (\bs
v_1,\cdots,\,\bs v_{N_c})\in\Rbb^{N_c\times N_c}$, with $\cl
B^{-1}=\cl B$.  The graph Laplacian eigenvector basis is the
generalization of the Fourier basis. For regular distribution of nodes
on an infinite plane, $\cl B$ coincides with the 2-D Fourier
basis. The Fourier transform of a vector $\bs f\in\Rbb^{N_c}$ living
on $\cl G$ is therefore naturally defined as
$$
\bs{\hat{f}} = \cl B^T \bs f,\quad \text{or}\quad \hat{f}_j = \bs v_j^T\,\bs f,\quad
\forall 1\leq j\leq N_c.
$$ 

Interestingly, this Graph Fourier Transform (GFT) is invariant under
any relabeling of the graph nodes, a useful property since there is no
reason why the salient points discovered by the corner detector should
be ordered similarly between two similar images. Indeed, given a
permutation matrix $\Pi\in\{0,1\}^{N_c\times N_c}$ with only one 1 per
row and column and $\Pi^{-1}=\Pi$, it is easy to show that if the
nodes of $\cl V$ are permuted accordingly, $\bs f \to \Pi \bs f$,
$\Delta \to \Pi\,\Delta\,\Pi^T$ and $ \bs{\hat{f}} \to (\Pi \cl B)^T
\Pi\, \bs f = \bs{\hat f}$. Thanks to this GFT, we propose the
following image hashing.

\begin{definition}[Invariant Spectral Hashing]
Given a certain \emph{Saliency Function} $\bs f\in\Rbb^{N_c}$ of $I$, namely
a function depending on the salient point locations and on the image intensity
$I$, the Invariant Spectral Graph (ISH) of $I$ is the spectrum of $\bs
f$, that is,
$$
\bs \varphi_{\rm Sp}(I)\ =\ |\bs{\hat f}(I)|\ \in\ \Rbb_+^{N_c}, 
$$
combined with the knowledge of the Saliency Graph Laplacian spectrum
$\{\lambda_i:1\leq i\leq N_c\}$.
\end{definition}
In this hash, the absolute value (applied component wise on the FT
vector) removes the ambiguity on the eigenvector
orientation\footnote{Laplacian eigenvector
  orientation is undetermined since $\Delta\,(\pm \bs
  v)=\lambda\,(\pm \bs v)$ for any eigenvector~$v$.}.  Consequently,
the ISH of $I$ contains information about both salient point
distribution (through the underlying graph) and image intensity
(through the saliency function).

\sq
\paragraph{Saliency function:} 
There exist of course an infinite choice of saliency
functions. Given the Saliency Graph $\cl G$ of an image $I$ determined
from $N_c$ salient points, we focus our approach on this one
$$
f_i = f(\bs c_i) = {\rm Var} \{I(\bs x):\ \forall \bs x\in\Rbb^2,\
\|\bs x - \bs c_i\| \leq \sigma\},\quad 1\leq i\leq N_c,
$$ 
the value $\sigma$ being the smoothed Harris detector radius.  In
other words, our saliency function $\bs f = (f_1,\,\cdots,f_{N_c})^T$
is interested in the variance of $I$ in a neighborhood of each salient
point. Taking the variance instead of for instance the mean gives the
same impact to all the salient points whatever their intensity. What
matters here is the variability of $I$ around these, that is, a
variation that is linked to the corner contrast.

\paragraph{ISH Complexity:} Given an image $I$ of $N$ pixels, the
computational complexity of the ISH evaluation is split as
follows. For the smoothed Harris detector, the complexity is $O(N \log
N)$ by performing fast convolution in the Fourier (FFT) domain. The time
consuming part of the graph definition is the connectivity
estimation. This one can be optimized from $O(N_c^2)$ to $O(N_c)$ by
using a geographical quadtree data structure of the nodes. The
Laplacian eigenvector/eigenvalue decomposition has a complexity of
$O(N_c^3)$, with computation time of about 0.01s for $N_c=100$ on a
standard laptop. The saliency function is roughly estimated in $O(N_c
N)$ computations but it could be optimized with a slight variation of
its definition (\eg using the precomputed cornerness). Finally, the
GFT of $\bs f$ has complexity $O(k N_c)$ if it is restricted to the
$k$ first Fourier coefficients.

\sq
\paragraph{Distance between ISH:} In general, for two different images
and two different saliency graphs, the two resulting Laplacian
eigenvalue systems do not necessarily match. Therefore, in order to
develop a consistent distance definition, for any image $I$ related to
the ISH $\bs \varphi$ and to the Laplacian spectrum
$\{\lambda_1,\cdots,\lambda_{N_c}\}$, we first consider the continuous
linear interpolation $\tilde{\varphi}:\Rbb\to\Rbb_+$ of the couples
$\{(\sqrt{\lambda_1},\varphi_1), \cdots,
(\sqrt{\lambda_{N_c}},\varphi_{N_c})\}$ such that
$\tilde\varphi(\sqrt{\lambda_i}\,)=\varphi_i$, where the square root
enforces the common Fourier reading of the spectrum\footnote{On the
  line, a Fourier mode of frequency $\omega$ is a Laplacian
  eigenvector with eigenvalue $\omega^2$.}. Then, for two images $I$
and $I'$, their ISH distance up to the $k^{\rm th}$ eigenvalue ($1\leq
k\leq N_c$) is defined as
$$
(\cl D_{\rm Sp}(I,I'))^2 = \int^{(\min(\lambda_k,\lambda_k'))^{1/2}}_{0}\ |\widetilde{\bs\varphi}_{{\rm
Sp}}(\omega) - \widetilde{\bs\varphi}'_{{\rm Sp}}(\omega)|^2\ \ud\omega.
\sq
$$

\paragraph{Distance between Laplacian spectra:}
Since Laplacian eigenvalues encode the saliency graph geometry
\cite{chung97}, it is worth to introduce a distance between two
Laplacian spectra. With the notations of the previous section
this distance reads
\sq
$$
(\cl D_{\rm \Delta}(I,I'))^2\ =\ \sum_{i=1}^{k}
|\lambda_i-\lambda'_i|^2.\sq
$$ 
We will observe in the Sec.~\ref{sec:results} that this distance
can improve the performance of a characterization by $\cl D_{\rm Sp}$.
Indeed, for similar visual contents, both $\cl D_{\rm Sp}$ and $\cl
D_{\Delta}$ should be low, and so should be their product $\cl D_{\rm Sp}\cl
D_{\Delta}$.

\sq
\paragraph{Ordered hash (OH):} Of course, there is another very simple
hash defined from any saliency function $\bs f=\bs f(I)$. This is the
\emph{ordered hash} $\bs \varphi_{\rm ord}(I)\ =\ |\bs f^*|\ \in
\Rbb_+^{N_c}$, obtained by reordering the values of $\bs f$ in a
vector $\bs f^*$ such that $|f^*_{i}|>|f^*_{i+1}|$ for any $1\leq i<
N_c$. The distance between two ordered hashes is then simply computed
as $(\cl D_{\rm ord}(I,I'))^2\ =\ \|\bs \varphi_{\rm ord}-
\bs\varphi'_{\rm ord}\|^2$. As explained later, the ordered hash has a
good efficiency but it requires to uses all the $N_c$ sorted values in
order to reach the same results than a ISH using only a fraction of
the frequencies.

\sq
\section{Experiments}
\sq
\label{sec:results}

Image hashing pursues two competing goals: \emph{robustness} and
\emph{discrimination}. In other words, the distance between hashes
should be low for similar images (whatever the considered
transformations) and high for different visual contents. Whether two
images are similar or not can therefore be decided by comparing the
distance between their hashes with a threshold value $\cl T>0$, that
is, given a certain distance $\cl D$, two images $I$ and $I'$ are
characterized as ``similar'' if $\cl D(I,I') < \cl T$ (\emph{positive}
test), and different else (\emph{negative} test).

In this paper, we do not focus on an optimal threshold selection for
the distances of interest. We rather evaluate the common True
Positive (TP), True Negative (TN), False Positive (FP) and False
Negative (FN) quantities for all possible $\cl T$. This procedure
allows us to estimate \emph{(i)} the Receiver Operating Characteristic
(ROC) curves that presents the \emph{sensitivity} of the test, or True
Positive Rate (${\rm TPR}(\cl T) = {\rm TP} / ({\rm TP} + {\rm FN})$), versus
the False Positives Rates (${\rm FPR}(\cl T) = {\rm FP} / ({\rm FP} + {\rm TN})$), and
\emph{(ii)}, the Area Under the Curve (AUC) of the ROC equals
to the probability that a random pair of similar images would be
assigned a lower distance than this of a random pair of distinct
images~\cite{fawcett06}. This AUC quantifies the discrimination and
robustness performance of the ROC curves.

\begin{figure}[ht]
  \centering
  \subfigure
  {
    \raisebox{-1mm}{\includegraphics[height=5cm]{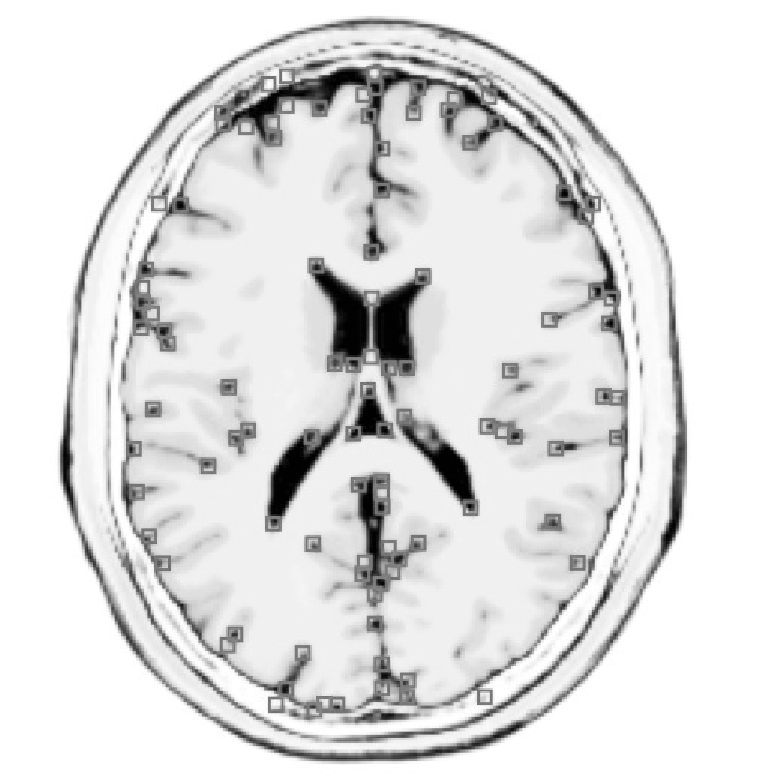}}
  }
  \subfigure
  {
    \includegraphics[height=5cm]{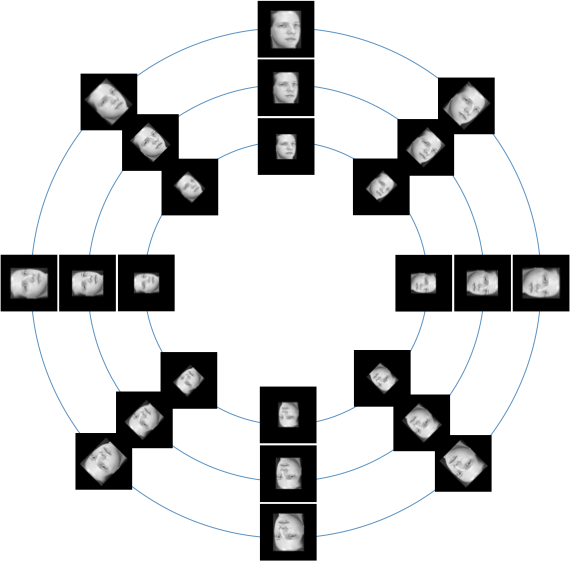}
  }
  \caption{(Left) Corners (gray squares) discovered by the smoothed Harris detector for
    a brain MRI image. (Right)~Schematic representation of the image
    manifold for the faces shown in polar coordinates ($\alpha,
    \theta$) where $\alpha$ is the scaling factor and $\theta$ is the
    rotation angle (Brain image manifold can be represented
    similarly). \label{fig:harris-example} \label{Image_manifold}}\sq
\end{figure}

\paragraph{Experimental setup:}
The databases used in our experiments are a T2-modulation volume MRI
cut into slices along the $xy-$directions, from the Brainweb
simulator~\cite{kwan99} and the ORL Database of
faces~\cite{samaria94}. In order to test the ISH, three sets of
transformations have been applied on these images: \emph{(i)} 9
rotations of angles between $0$ and $\pi$, \emph{(ii)} scalings of
factors between $0.8$ and $1.2$, \emph{(iii)} and 9 random
combinations of these rotations and scalings.  A schematic
illustration of the face image manifold (that has a polar
representation for each image) is shown in Fig.~\ref{Image_manifold}.

For each image, the number of extracted salient points was
set\footnote{When the number of salient points was smaller than $N_c$,
  the distances $\cl D_{\rm Sp}$, $\cl D_{\rm ord}$, and $\cl
  D_{\Delta}$ have been computed relatively to the smallest hash
  size.} to $N_c=100$ maximum, the Gaussian kernel used for saliency
detection has a standard deviation $\sigma$ of 2.5\% of the graph
diameter $d^*$. For the value of the connectivity radius $r$
(Sec.~\ref{sec:graph-definition}), good results have been obtained if
$r=1/15$. With this value, each node in the resulting saliency graphs
were connected to an average of 5 other nodes.

\sq
\paragraph{Results and discussions:} 

\begin{figure}[ht]
  \centering
  \sq
  \subfigure
  {
    \includegraphics[width=6cm]{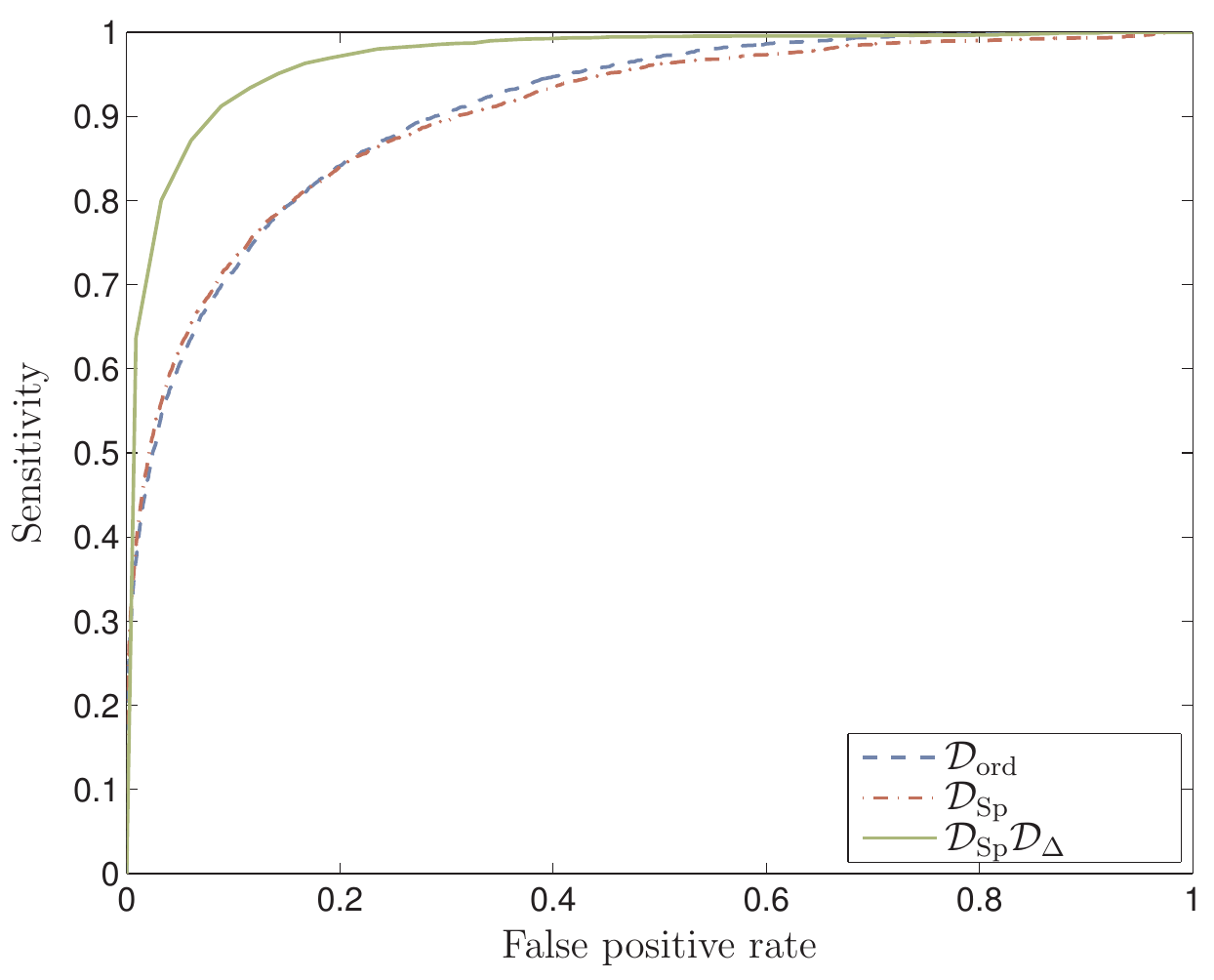}
  }
  \subfigure
  {
    \includegraphics[width=6cm]{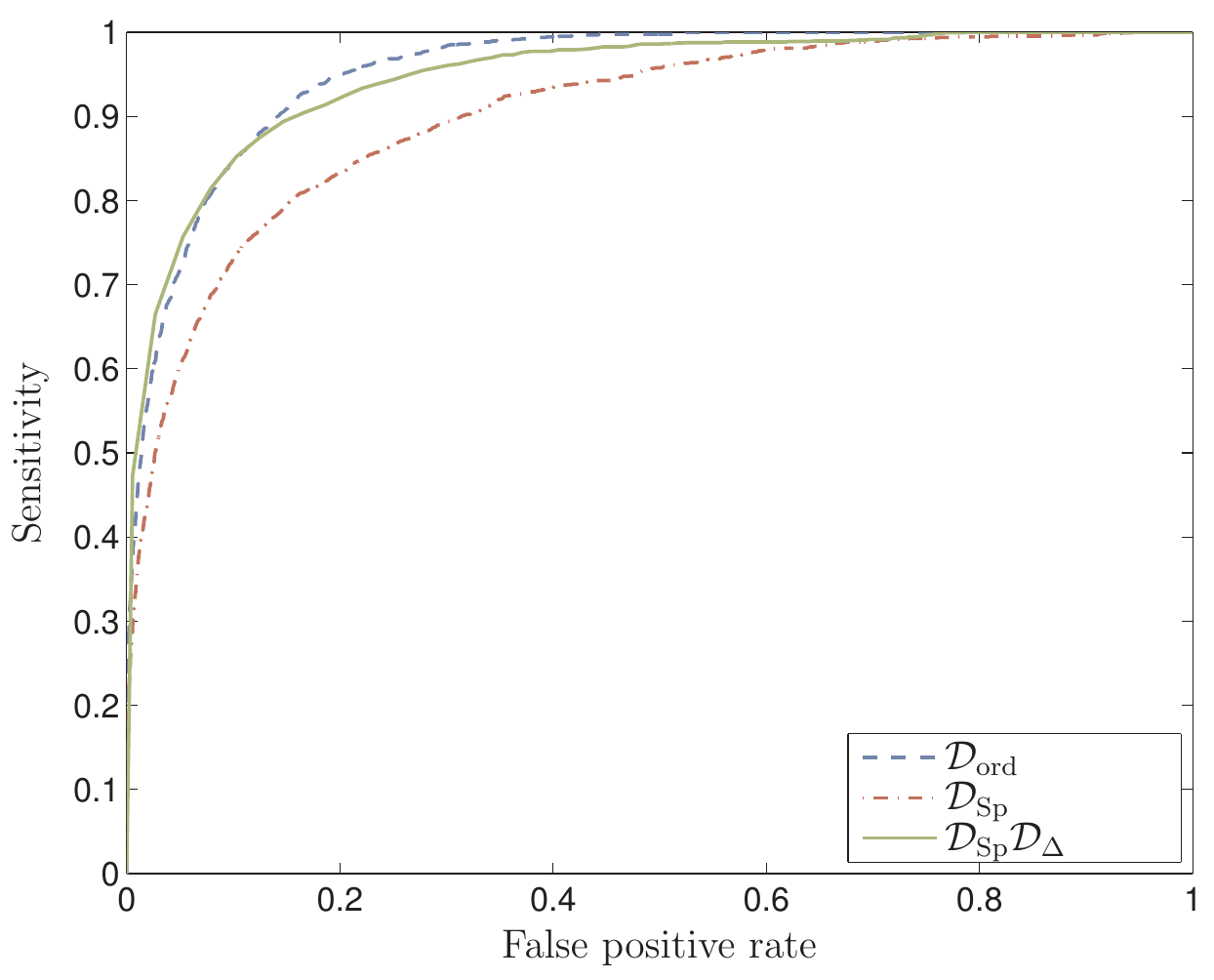}
  }
  \caption{ROC curves for mixed image rotation and scaling: (left) brain
    database, (right) faces database. The ``$-\cdot-$'', dashed and
    continuous curves show the robustness when $\cl D_{\rm ord}$, $\cl
    D_{\rm Sp}$ and $\cl D_{\rm Sp}\cl D_{\Delta}$ are used,
    respectively.}\sq
  \label{ROC-curves}
\end{figure}

\begin{figure}[ht]
  \begin{center}
    \subfigure
    {
      \footnotesize
      \begin{tabular}{|l|l|c|c|c|}
        \hline
        \textbf{Database}&\textbf{Transf.}&${\cl D_{\rm ord}}$&${\cl D_{\rm Sp}}$&${\cl D_{\rm Sp}\cl D_{\Delta}}$\\
        \hline
	         &Rotation     &0.695   &0.926&\bf 0.985\\
        Brainweb &Scaling      &0.724   &0.888&\bf 0.967\\
                 &Rot. \& Scal.&0.787   &0.904&\bf 0.969\\
        \hline
                 &Rotation     &0.892   &0.918&\bf 0.961\\
        ORL Faces&Scaling      &0.933   &0.891&\bf 0.938\\
                 &Rot. \& Scal.&\bf 0.954&0.902&0.946\\
        \hline
      \end{tabular}
    }
    \subfigure
    {
      \raisebox{-2.5cm}{\includegraphics[width=6.4cm]{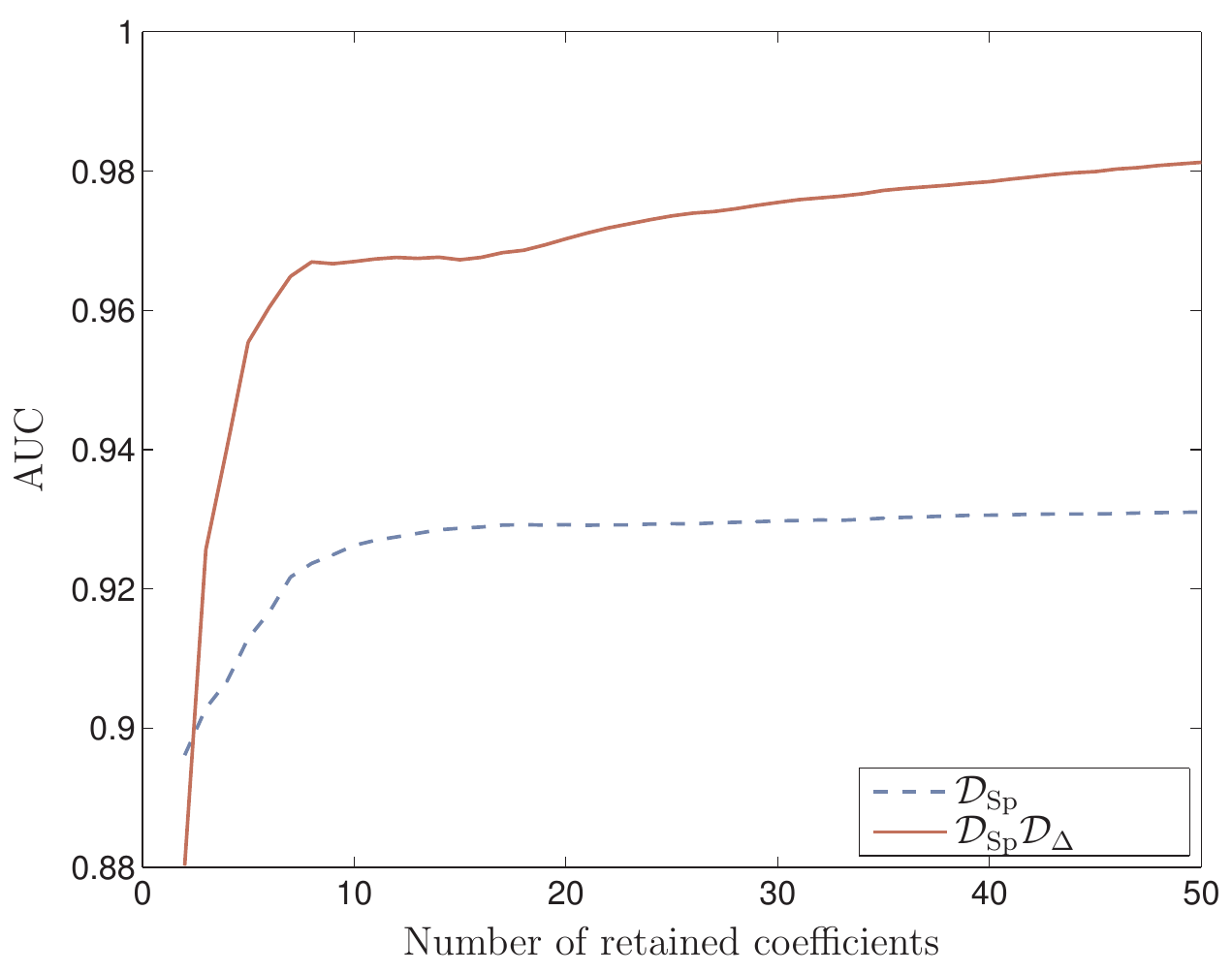}}
    }
  \caption{(Left) AUC table for the different transformations of the images
    using distance between ordered hashing ($\cl D_{\rm ord}$),
    distance between spectral hashing ($\cl D_{\rm Sp}$), and spectral
    hashing combined with spectra comparison ($\cl D_{\rm Sp}\cl
    D_\Delta$). (Right)~Area under the ROC curve for increasing number of
    ISH coefficients and eigenvalues. Most information is contained in
    the 10 first coefficients after what the AUC remains mostly
    constant. \label{AUC-table} \label{AUC}}\sq
\end{center}
\end{figure}

The ROC curves testing rotation invariance, scaling invariance, and
mixed rotation and scaling invariance have been computed for the two
databases and for $\cl D_{\rm ord}$, $\cl D_{\rm Sp}$ and $\cl D_{\rm
  Sp} \cl D_{\Delta}$. For these two last distances, the ROC curves
have been obtained by keeping only the $k=10$ first eigenvalues and
GFT coefficients. The ROC curves testing mixed rotation and scaling are
shown in Fig.~\ref{ROC-curves} for the two databases. All the
related AUCs are summarized in~Fig.~\ref{AUC-table}.

For the brain MRI database, $\cl D_{\rm Sp}$ achieves sensitivities
over 90\% with false positives rates lower than 10\%. Under rotation
only, a sensitivity of 95\% with a false positives rate of 8\% is
achieved. Results for the ORL Database of Faces were slightly worse
due to the lower number of salient points detected. For all faces, the
maximum possible number of salient points, \ie all the local maxima of
the cornerness function, was systematically lower than the imposed
maximum of $N_c=100$. The hashing was therefore more sensitive to the
variations of salient point positions between different transformation of
the same image.

\begin{figure}[t!]
  \centering
  \subfigure
  {\includegraphics[width=6cm]{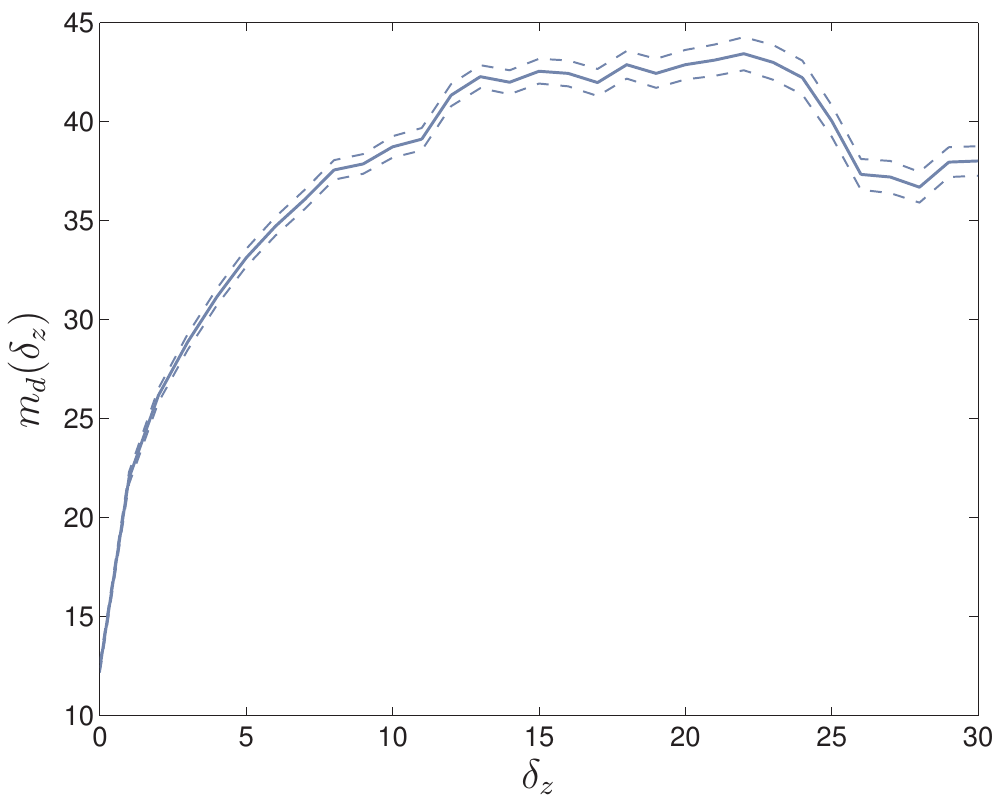}}
  \subfigure
  {\raisebox{2mm}{\includegraphics[width=7cm]{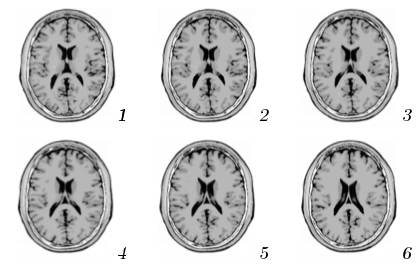}}}
  \caption{(Left) The expectation of the distance between slices of the brain
    database is an increasing function of the actual physical distance
    between slices. (Right)~Six contiguous slices of the brain. The visual content
    does not change much from slice to slice. A suitable image metric
    should therefore yield low distances between
    them. \label{mdDeltaz} \label{BrainSlices}}\sq\sq 
\end{figure}

For the brain database, an interesting result is that the distance
between brain slices that are physically close is shorter than the
distance between slices wide apart in the brain. In other words, if we
consider the brain MRI as a volumetric image
$I_z(\bs x)=I(x,y,z)$ for which each slice in the database is the
result of fixing $z$ to some value, then the expectation of the
distance $\cl D_{\rm Sp}$ between two slices separated along the $z-$axis by
a distance $\delta_z>0$,
\begin{equation}
m_d(\delta_z) = \Ebb\big\{ \cl D_{\rm Sp}(I_z,I_{z+\delta}): \delta =
\pm \delta_z, z\in\Rbb\big\},
\end{equation}
is an increasing function of $\delta_z$ for $\delta_z$ sufficiently
close to zero. Fig.~\ref{mdDeltaz} shows the evolution of
$m_d(\delta_z)$ (with two dashed curves providing the 99\% confidence
interval on $m_d$ estimation) computed over the 100 brain slices
rotated and scaled. The expectation of the distance is indeed
increasing for $\delta_z\leq 10$. This means that the distance $\cl
D_{\rm Sp}$ between ISH truly reflects the difference between visual
contents. The MRI was indeed taken with a $z-$resolution of 1mm for
which visual contents of contiguous slices are very close, as depicted
in Fig.~\ref{BrainSlices}. The false positive pairs of images are
therefore more likely to be adjacent slides which are visually close
than totally different slides.

In order to validate the performance of the ISH compared to the naive
ordered hashing (OR) ordering the $N_c$ values of the saliency
function, all the ROC curves were computed using only the $k=10$ first
GFT coefficients and the 10 first Laplacian eigenvalues.  Therefore,
the hash lengths related to the use of $\cl D_{\rm Sp}$ and $\cl
D_{\rm Sp}\cl D_{\Delta}$ are both equal to 20, that is, 20\% of the
tested OR hash length. Results show that the spectral hashing with
less coefficients performs as good or better than the naive ordering
hashing. It is interesting to quantify the gain in discrimination when
the number $k$ of GFT coefficients and eigenvalues increases in the
spectral hashing. This can be evaluated by computing the AUC for an
increasing number of coefficients. This evolution is depicted in
Fig.~\ref{AUC} for both the spectral hashing and the combination of
the spectral hashing with the spectral comparison. As a result, the
performance does not increase much when more than 2$\times$10
coefficients are retained. The spectral hashing is therefore capable
to extract the information useful to discriminate between different
visual contents in fewer coefficients.
 
\sq
\section{Conclusion}
\sq
\label{sec:conclusion}

This paper has shown that the geometry of salient point distribution
can advantageously be considered in order to form an invariant image
hashing. This geometrical inclusion is achieved through the Laplacian
spectrum of a Saliency Graph built by connecting geographically close
salient points. In consequence, the associated Graph Fourier Transform
of some saliency function, that can be improved with the Laplacian
eigenvalue distribution, provides a robust and discriminant image
hashing. Moreover, compared to the ordered hashing where the knowledge
of the salient point distribution is lost, the Invariant Spectral
Hashing requires much less values for the same efficiency. In a future
research, the impact of the connectivity parameters (like the radius
$r$) on the classification procedure will be assessed, together with a
careful study of different quantization strategies (\eg scalar
quantization of the different spectra). We expect also to achieve a
characterization of images made of several distinct objects arranged
on a smooth background. The saliency graph can indeed serve to
partition the image thanks to the structure of the first Laplacian
eigenvectors (like the zero crossing paths).

\sq
\footnotesize

\end{document}